\def\paperTitle{PLOT: Text-based Person Search with Part Slot Attention for Corresponding Part Discovery
}

\def\abbvPaperTitle{PLOT: TBPS with Part Slot Attention for Corresponding Part Discovery}

\def\paperTitleSupp{PLOT: Text-based Person Search with Part Slot Attention for Corresponding Part Discovery \texorpdfstring{\\} \emph{--- Supplementary Materials ---}}

\def\authorBlock{
    Jicheol Park \orcidlink{0009-0004-7899-6802} \and
    Dongwon Kim \orcidlink{0000-0003-1147-5274} \and
    Boseung Jeong \orcidlink{0000-0001-9382-3396} \and
    Suha Kwak \orcidlink{0000-0002-4567-9091}
    }

\def\authorInstitutes{
    Pohang University of Science and Technology (POSTECH), South Korea\\
    \email{\{jicheol, kdwon, boseung01, suha.kwak\}@posetech.ac.kr}
    \url{https://cvlab.postech.ac.kr/research/PLOT}
}

\def\abbvAuthor{J.~Park et al.}

\newif\ifreview 
\newif\ifcamera 
\newif\ifmobile 
\documentclass[runningheads]{llncs}

\usepackage{eccv}
 
\usepackage{eccvabbrv}

\usepackage{graphicx}
\usepackage{amsmath}	
\usepackage{amssymb}	
\usepackage{booktabs}
\usepackage{microtype}
\usepackage{epsfig}
\usepackage{caption}
\usepackage{float}
\usepackage{placeins}
\usepackage{color, colortbl}
\usepackage{stfloats}
\usepackage{enumitem}
\usepackage{tabularx}
\usepackage{xstring}
\usepackage{multirow}
\usepackage{xspace}
\usepackage{url}
\usepackage{xcolor}
\usepackage[hang,flushmargin]{footmisc}
\usepackage{pifont}
\usepackage{wrapfig}
\usepackage{lmodern}

\usepackage{siunitx}
\usepackage{newtxmath}

\usepackage[accsupp]{axessibility}  

\newcommand{\cmark}{\ding{51}}
\newcommand{\xmark}{\ding{55}}

\def\eg{\emph{e.g.}}
\def\etal{\emph{et al.}}

\definecolor{grn}{rgb}{0.1, 0.6, 0.1}
\definecolor{mgt}{rgb}{0.6, 0.1, 0.6}
\definecolor{red}{rgb}{0.7 0.1 0.1}
\definecolor{blue}{rgb}{0.0 0.0 1.0}
\definecolor{amethyst}{rgb}{0.6, 0.4, 0.8}
\definecolor{Gray}{gray}{0.9}
\definecolor{grey}{rgb}{0.9, 0.9, 0.9}
\newcommand{\ccol}{\cellcolor{grey}}
\newcommand{\expnum}[2]{{#1}\mathrm{e}{-#2}}

\newcommand{\sftype}[1]{{\textsf{\small #1}}}

\newcommand{\dw}[1]{{\color{black}{#1}}}

\usepackage[breaklinks,colorlinks,citecolor=eccvblue]{hyperref}

\usepackage{orcidlink}

\begin{document}

\title{\paperTitle}

\titlerunning{\abbvPaperTitle}

\author{\authorBlock}
\authorrunning{\abbvAuthor}
\institute{\authorInstitutes}

\maketitle

\begin{abstract}
Text-based person search, employing free-form text queries to identify individuals within a vast image collection, presents a unique challenge in aligning visual and textual representations, particularly at the human part level. 
Existing methods often
struggle with part feature extraction and alignment due to the lack of direct part-level supervision and reliance on heuristic features. 
We propose a novel framework that leverages a part discovery module based on slot attention to autonomously identify and align distinctive parts across modalities, enhancing interpretability and retrieval accuracy without explicit part-level correspondence supervision. 
Additionally, text-based dynamic part attention adjusts the importance of each part, further improving retrieval outcomes.  
Our method is evaluated on three public benchmarks, significantly outperforming existing methods. 

\end{abstract}
\section{Introduction}
\label{sec:intro}

Text-based person search is the task of identifying the target person from the vast collection of images with a free-form text query.
This task demands extracting identifiable features, such as human parts, from both textual and visual modalities to capture subtle differences between individuals.
Hence, establishing correspondence between the extracted human part features across image and text modalities is essential for accurate text-based person search.
However, it is not straightforward to extract these part features and establish their correspondences between the two modalities without part-level supervision.

\begin{figure*}[t!]
    \centering
    \includegraphics[width=\textwidth]{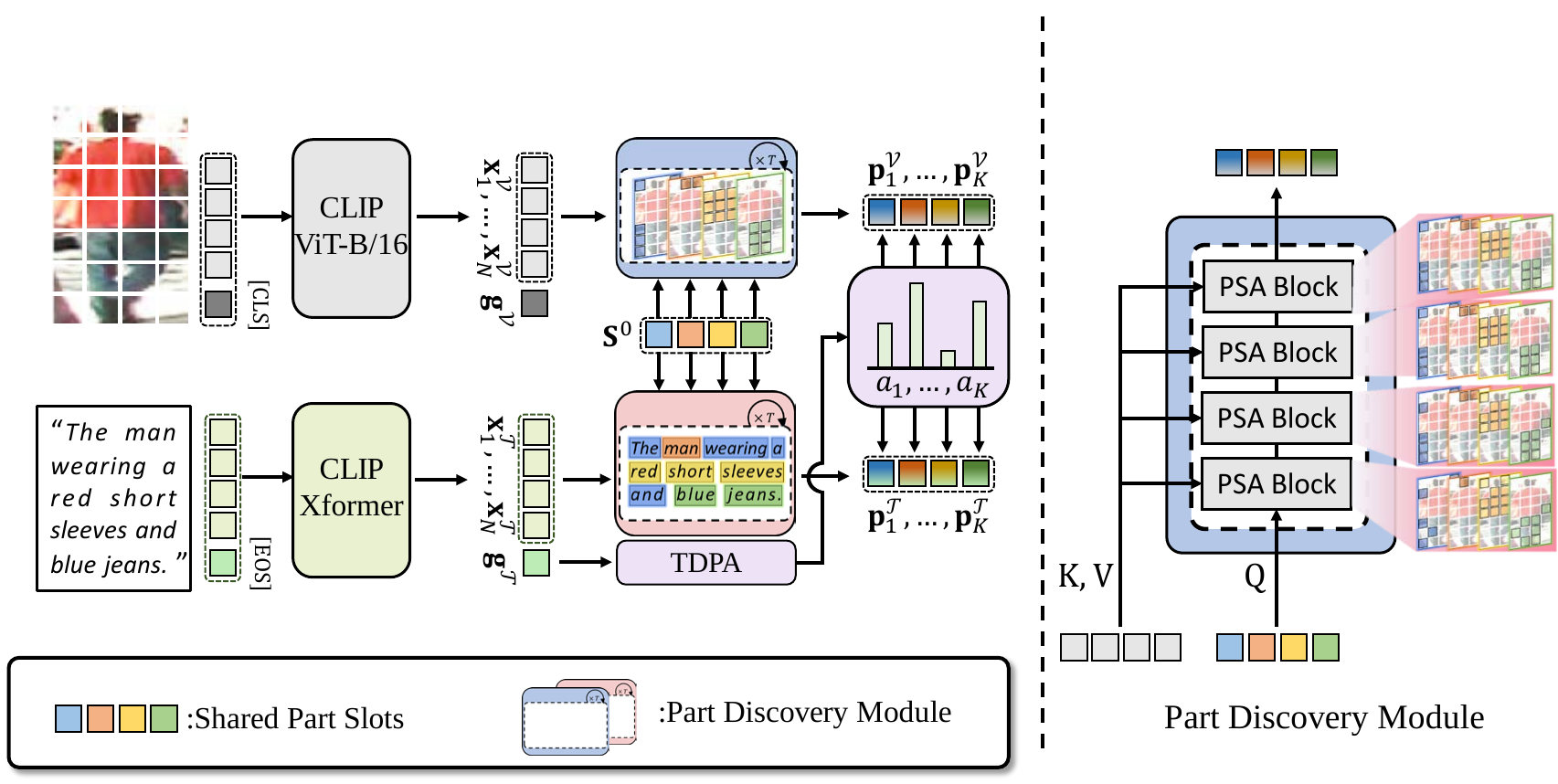}
    \vspace{-6mm}
    \caption{
    The overall architecture of PLOT. 
    }
    \label{fig:architecture}
    \vspace{-5mm}
\end{figure*}

To address this challenge, previous work~\cite{chen2022tipcb, suo2022simple} relies on the heuristic part features obtained by equi-horizontally cropping the entire image;
such features are then matched to the free-form text queries for person search.
However, the heuristic part features used in this approach are susceptible to deformation caused by occlusion and pose variation.
Meanwhile, earlier work~\cite{jing2020pose, wang2020vitaa, lgur} proposed a learning-based approach for part feature extraction.
Nevertheless, these methods tend to generate redundant part features that lack disentanglement~\cite{lgur}, or they demand access to additional part-level supervisions~\cite{jing2020pose, wang2020vitaa}.

To tackle the above issues, we introduce a new framework that discovers distinctive parts in both modalities and matches them between the two modalities without any correspondence supervision; its overall architecture is illustrated in Fig.~\ref{fig:architecture}.
To discover distinctive parts and extract their features from both modalities, we propose PLOT, a \textbf{P}art discovery module based on the s\textbf{LOT} attention mechanism~\cite{slot}.
Slot attention is an attention mechanism designed for object-centric learning, which segments input data into a set of slots representing individual entities without requiring object-level supervision. 
In PLOT, we first define a set of learnable embedding vectors, termed \emph{part slots}, that contain primitive information related to human body parts shared between the two modalities.
Then, these part slots undergo refinement through several iterative attention processes, where they compete amongst themselves to bind with the input data; ultimately, the part slots are transformed into part features, termed \emph{part embeddings}, that represent distinctive parts in the input data.

To ensure correspondence between part embeddings from different modalities, the part slots are shared between the visual and textual modalities.
Part embeddings from the same part slots are then learned to represent the identical part although they are computed from different modalities.
This mechanism enables PLOT to match the discovered parts from the two modalities without supervision for the correspondence as well as capturing part-level fine-grained appearance features from the both modalities.
Hence, PLOT improves performance of text-based person search through the rich and fine-grained part features, and at the same time, it guarantees interpretable retrieval by providing part-level correspondences between query text and retrieved images.

In addition, PLOT introduces a new similarity aggregation method for part embeddings called text-based dynamic part attention (TDPA) pooling, which dynamically adjusts the weights of part embeddings based on the text query.
Given a text query as input, TDPA predicts the importance weight of each slot for retrieval and applies the predicted importance weights to aggregate the similarities of part embeddings between the two modalities.
TDPA allows the entire retrieval system to perform optimized retrieval for each query, leading to improved performance.

Our method was evaluated and compared with prior work on three public benchmarks~\cite{li2017person, ding2021semantically, zhu2021dssl}, where it clearly outperformed all existing methods thanks to the rich representation based on part embeddings.
The main contribution of our work is four-fold:
\begin{itemize}
   \vspace{-1mm}\item We introduce PLOT, a new framework for text-based person search that discovers distinguished human body parts, extracts their embeddings, and establishes their correspondences between the two modalities with no human intervention.
   \item The part embeddings provided by PLOT enables an interpretable text-based person search thanks to the part-level correspondences it provides.
   \item We introduce a novel similarity aggregation method that adaptively determines importance of each discovered part based on each text query and consequently enables retrieval optimized per query.
   \item Our model with PLOT achieved the best on all the public benchmarks for text-based person search.
\end{itemize}



\section{Related Work}
\label{sec:related}

\subsection{Text-Based Person Search}
In recent years, the task of text-based person search has gained significant attention in the computer vision community. 
Li~\etal,~\cite{li2017person} proposed a gated neural attention-based recurrent neural network (GNA-RNN) for learning the affinity between text descriptions and images, along with providing a benchmark dataset CUHK-PEDES for model evaluation.
Zhang~\etal,~\cite{zhang2018deep} proposed cross-modal projection matching and classification (CMPM+CMPC) loss, for learning deep discriminative image-text global embeddings.
However, these methods primarily focus on global representations of input data and are thus not capable of capturing distinctive human part details, leading to limited performance in the text-based person search.

To address the above problem, a line of studies focuses on extracting fine-grained representations.
One of the prominent examples of exploiting fine-grained information is to cut human images horizontally and use them as human parts~\cite{gao2021contextual, niu2020improving, wang2021text, chen2022tipcb, ding2021semantically, suo2022simple}.
Chen~\etal,~\cite{chen2022tipcb} extracts image part embeddings through equi-horizontal cropping of the entire image and aligns additional network to transform the textual global into corresponding parts. However, these heuristic part features inevitably include no-informative information, such as background elements.
To avoid the above limitations, Suo~\etal,~\cite{suo2022simple} proposed the simple and robust correlation filtering method to extract foreground features on the heuristic part features.
Yet, the heuristic divided part features fundamentally fall short of capturing the complex human parts, underscoring a critical limitation in their expressive capability.
To move beyond such heuristic part structures, Shao~\etal,~\cite{chen2022tipcb} proposes a learning-based approach for part feature extraction.
However, this method struggles to extract distinctive part embeddings due to the extraction of redundant part information, which lacks sufficient disentanglement.
To extract exquisite human parts, several studies tried to utilize useful information (\eg, human attributes and human keypoints) via external tools~\cite{aggarwal2020text, wang2020vitaa, jing2020pose}.
Wang~\etal,~\cite{wang2020vitaa} introduced an auxiliary attribute segmentation to align the visual part features with the textual attributes parsed from text description.
Jing~\etal,~\cite{jing2020pose} proposed a new multi-granularity attention network to learn the part feature alignment between visual and textual with human pose estimation. 
However, these approaches have inevitable limitations of high computational cost and dependence on the performance of external tools for local feature extraction.

Most recent work, Jiang~\etal,~\cite{irra} utilizes pre-trained CLIP~\cite{clip} model, capitalizing on the rich knowledge of models trained on extensive data for text-to-image matching, to excel in text-based person search.
Yet, this methods primarily focus on global features, not specifically designed for extracting human part features.
Not only do we leverage pre-trained knowledge, but we transcend heuristic human parts methods to extract sophisticated human parts without external tools or part-level supervision.



\subsection{Slot Attention}
%
\dw{Slot attention~\cite{slot} is a recently proposed attention mechanism for object-centric learning, a problem focusing on discovering constituent visual entities within an image.
The unique property of slot attention is that it can represent input images as a set of slots, where the slots are representations corresponding to individual visual entities, without any object-level supervision during training.
Within slot attention, slots iteratively compete for aggregating input data, ensuring distinct representations focusing on individual visual entities.
By incorporating slot attention, the proposed framework facilitates the unsupervised identification of a structure underlying image and text queries, enabling the model to discern and represent individual human parts without explicit supervision. 
This capability is particularly valuable in person search datasets, where recognizing subtle differences and understanding the correspondence of human parts across different modalities are crucial. }
\section{Proposed Method}
\label{sec:method}

\dw{
The following subsection offers details of global and part embeddings extraction for each modality with its backbone (Sec ~\ref{sec:embedding_extraction}).
Subsequently, we present our novel framework, PLOT, which includes the part discovery module (Sec.~\ref{sec:part_feature}) and similarity aggregation between part embeddings (Sec.~\ref{sec:similarity}),
concluding with a discussion on the learning objective designed to optimize our proposed framework (Sec.~\ref{sec:loss}) and inference of our framework (Sec.~\ref{sec:inference}).
}

\subsection{Global and Part Embeddings}\label{sec:embedding_extraction}
In our framework, a single input data is described by two different types of representations: a global embedding and multiple part embeddings.
The global embedding is used to represent the input data holistically, while each part embedding describes the appearance of distinctive human parts (\eg,~arm, leg, torso, etc.).
We below provide details of global and part embeddings computation for visual and textual modalities.
Following the previous work~\cite{irra}, a pre-trained CLIP~\cite{clip} is used as the backbone networks for the visual and textual modalities. 

\vspace{1mm}\noindent\textbf{Visual Modality:}
We utilize the vision transformer (ViT) from the CLIP-B/16~\cite{clip} architecture as a visual backbone network. 
Initially, an input image of a person is split into $N$ distinct, non-overlapping patches, which are subsequently transformed into patch tokens through linear projection.
The patch tokens and an extra \texttt{[cls]} token are then fed into the visual backbone network.
The token sequence is processed throughout multiple self-attention blocks, and the \texttt{[cls]} token of the last block is used as a global embedding $\mathbf{g}^\mathcal{V} \in \mathbb{R}^{D}$.
To obtain part embeddings $\mathbf{P}^\mathcal{V} \in \mathbb{R}^{K \times D}$, remaining patch tokens of last block $\mathbf{x}^\mathcal{V} \in \mathbb{R}^{N \times D}$ is passed to the part discovery module.
The part discovery module aggregates the patch features describing coherent human parts into the same part embedding.
We provide more detailed information about the part discovery module in Sec.~\ref{sec:part_feature}.

\vspace{1mm}\noindent\textbf{Textual Modality:}
For the textual backbone network, we utilize a transformer architecture from the CLIP-Xformer~\cite{clip} text encoder. This encoder operates on text input transformed into byte pair encoding (BPE) sequences. 
Initially, the text query undergoes tokenization via BPE, followed by the enclosing with \texttt{[SOS]} and \texttt{[EOS]} tokens. 
The resulting sequence of tokens is then inputted into the textual backbone network. 
Here, the \texttt{[EOS]} token from the final block serves as the global embedding, denoted as $\mathbf{g}^\mathcal{T} \in \mathbb{R}^{D}$. 
Analogous to the approach for the visual modality, we process the remainder of the text tokens, $\mathbf{x}^\mathcal{T} \in \mathbb{R}^{L \times D}$, into part embeddings, $\mathbf{P}^\mathcal{T} \in \mathbb{R}^{K \times D}$, utilizing the part discovery module.

\subsection{Part Discovery Module}\label{sec:part_feature}
For extracting part embeddings in each modality, part discovery module aggregates patch tokens $\textbf{x}^{\mathcal{V}}$ and text tokens $\textbf{x}^{\mathcal{T}}$ into the visual part embeddings $\mathbf{P}^\mathcal{V}$ and textual part embeddings $\mathbf{P}^\mathcal{T}$, respectively.
It is worth noting that the part discovery module for each modality has identical model architecture and functions equivalently.
Therefore, we will only explain the part discovery module on the visual modality for brevity.


Part discovery module consists of initial part slots $\mathbf{S}^0\in \mathbb{R}^{K\times D}$ and $T$ multiple iteration of the \textit{part slot attention block} (PSA block).
To extract part embeddings, we first initialize a set of learnable embeddings \emph{part slots} $\mathbf{S}^0\in \mathbb{R}^{K\times D}$, where $K$ indicates the number of part slots.
Then, through a series of $T$ iterations of our PSA block, the initial $\mathbf{S}^0$ evolves into refined $\mathbf{S}^T$, where each slot captures distinct parts within input data.
The refined part slots $\mathbf{S}^T$ are used as visual part embeddings $\mathbf{P}^\mathcal{V}$.
The part discovery module can be formulated as follows:
\begin{align}
    \mathbf{P}^\mathcal{V} := \mathbf{S}^T, \text{where } ~\mathbf{S}^t = \texttt{PSA\_Block}^\mathcal{V}(\textbf{x}^{\mathcal{V}};\mathbf{S}^{t-1}).
    \label{eq:part_emb}
\end{align}

The PSA block first transforms the inputs $\mathbf{S}^{t-1}$ and $\textbf{x}^{\mathcal{V}}$ with layer normalization and linear projection layers $q(\cdot),k(\cdot)$ and $v(\cdot)$ to obtain embeddings of $D_h$ dimension.
Then the attention map $A \in \mathbb{R}^{N\times K}$ between $\mathbf{S}^{t-1}$ and $\textbf{x}^{\mathcal{V}}$ is computed by
 \begin{equation}
    A_{n,k}=\frac{e^{M_{n,k}}}{\sum_{i=1}^{K}e^{M_{n,i}}},~ \textrm{where} ~M = \frac{k(\textbf{x}^{\mathcal{V}})q(\mathbf{S}^{t-1})^\top}{\sqrt{D_h}}.
    \label{eq:part_slot_attention}
 \end{equation}
The attention map is obtained through normalization across part slots; this normalization encourages competition among themselves to bind distinct sets of patch tokens to each slot.
We update part slots with weighted mean of patch tokens obtained by attention map $A_{n,k}$ and then feed it to a gated recurrent unit (GRU) using $\mathbf{S}^{t-1}$ as hidden state as follows:
 \begin{equation}
    \bar{A}_{n,k}=\frac{A_{n,k}}{\sum_{i=1}^{N}{A_{i,k}}},~ \bar{\mathbf{S}}^t =\texttt{GRU}\big(\mathbf{S}^{t-1},\bar{A}^{\top}v(\textbf{x}^{\mathcal{V}})\big).
    \label{eq:update_slot_attention}
 \end{equation}
Then, we obtain $t$-th part slot by feeding $\bar{\mathbf{S}}^{t-1}$ into a multi-layer perceptron (MLP) with layer normalization, ReLU activation, and residual connection:
\begin{equation}
    \mathbf{S}^t = \texttt{PSA\_Block}^\mathcal{V}(\textbf{x}^{\mathcal{V}};\mathbf{S}^{t-1}) = \texttt{MLP}(\bar{\mathbf{S}}^{t-1}) + \bar{\mathbf{S}}^{t-1}.
    \label{eq:part_slot}
 \end{equation}
Finally, we can obtain the visual part embeddings $\textbf{P}^{\mathcal{V}} = \{\mathbf{p}_k^{\mathcal{V}}\}^K_{k=1}$ which is the output of $T$-th iteration of PSA block: $\textbf{P}^{\mathcal{V}} = \mathbf{S}^T_\mathcal{V} \in \mathbb{R}^{K\times D}$.

\vspace{1mm}\noindent\textbf{Part Correspondence through Slot Sharing:}
Additionally, we share the learnable part slots $\mathbf{S}^0$ between two part discovery modules to establish correspondences between part embeddings extracted from each modality, considering that part emeddings extracted from the same part slot are corresponding part across modalities, thereby contributing to a clearer comparison between modalities.

\subsection{Measuring Similarity between Embeddings}\label{sec:similarity}

\dw{
In the context of training and applying our retrieval model, selecting an appropriate similarity function between embeddings is crucial. 
The challenge lies in dealing with two distinct types of embeddings: global embeddings and part embeddings. 
For global embeddings, cosine similarity offers a straightforward and effective means of measuring similarity.

However, the situation becomes more complex when considering part embeddings.
A direct method for addressing this complexity involves calculating the average cosine similarity across all pairs of part embeddings. 
This approach, while straightforward, has its drawbacks, primarily because it treats all part-wise similarities as equally significant.
In reality, the relevance of specific part embeddings to the actual similarity between data instances can significantly vary, influenced by the context of a text query.
For instance, if a text query focuses exclusively on particular features of a human figure, the similarity contributions from other unrelated part embeddings should be less relevant.  
This challenge highlights the necessity for an approach that can dynamically assess and prioritize the relevance of part embeddings based on the context provided by the query.

To address this challenge, we introduce text-based dynamic part attention (TDPA) to aggregate the similarities between part embeddings.}
Firstly, textual global embedding $\textbf{g}^{\mathcal{T}}$ is transformed to the TDPA $\mathbf{a} \in \mathbb{R}^{K}$, using MLP and a softmax function.
Using TDPA, the aggregated similarity between part embeddings are computed as follows:
 \begin{align}
    &\mathbf{a} = \sigma\big(\texttt{MLP}(\textbf{g}^{\mathcal{T}})\big) \in \mathbb{R}^{K}, \\
    &c_\text{agg}(\textbf{P}^\mathcal{V},\textbf{P}^{\mathcal{T}}; \textbf{g}^{\mathcal{T}}) := \sum_{k=1}^{K} \mathrm{a}_k\cdot c(\mathbf{p}^{\mathcal{V}}_k,\mathbf{p}^{\mathcal{T}}_k),
    \label{eq: part similarity}
 \end{align}
where $c(\cdot, \cdot)$ denotes cosine similarity between two embeddings, $\mathrm{a}_k$ is the $k$-th value of $\mathbf{a}$, and $\sigma(\cdot)$ is a softmax function. 
If given textual global embedding $\textbf{g}^{\mathcal{T}}$, TDPA is computed by \texttt{MLP} which is learned in an end-to-end manner by minimizing our partNCE loss, as will be introduced in Sec~\ref{sec:loss}.
This dynamic attention enables us to adaptively assign importance to each part embedding, which reduces the impact of non-informative part embedding and promotes a more informative similarity measurement that reflects the significant semantic similarity of each text query.

\subsection{Learning Objective}\label{sec:loss}
Our model is trained through the establishment of cross-modal alignments, considering both global and part levels.
The alignment at the global-level is accomplished by aligning global embeddings $\textbf{g}^{\mathcal{V}}$ and $\textbf{g}^{\mathcal{T}}$ that capture the comprehensive information of each modality.
For part-level alignment, we leverage part embeddings $\textbf{P}^{\mathcal{V}}$ and $\textbf{P}^{\mathcal{T}}$ that have locally exclusive features within each modality due to the slot attention. 
Furthermore, by dynamically adjusting weights of specific part slots based on the text global embedding $\textbf{g}^{\mathcal{T}}$, we facilitate the learning of more informative alignments.

\vspace{1mm}\noindent\textbf{Global Alignment Loss:}
To align global embeddings $\textbf{g}^{\mathcal{V}}$ and $\textbf{g}^{\mathcal{T}}$ extracted from each modality backbone, we first define a batch of global embeddings $\mathcal{B}_\text{global} = \{(\textbf{g}^{\mathcal{V}}_i, \textbf{g}^{\mathcal{T}}_i)\}_{i = 1}^B$, where $B$ is batch size.
Then we adopt the InfoNCE loss~\cite{infonce} which is a contrastive learning objective that maximizes the similarity between embeddings of positive pairs while minimizing the similarity between negative pairs in the batch.
Consequently, our globalNCE loss with cosine similarity function $c(\cdot, \cdot)$ is formulated as follows:
\begin{equation}
    \mathcal{L}_{\textrm{NCE}}= -\sum_{i=1}^{B}\left(\textrm{log}
    \frac{e^{c(\textbf{g}^{\mathcal{V}}_i,\textbf{g}^{\mathcal{T}}_{i}) /\tau}}
        {\sum_{j=1}^B e^{c(\textbf{g}^{\mathcal{V}}_i,\textbf{g}^{\mathcal{T}}_j) /\tau}}
        + 
    \textrm{log}
    \frac{e^{c(\textbf{g}^{\mathcal{V}}_i,\textbf{g}^{\mathcal{T}}_{i}) /\tau}}
        {\sum_{j=1}^B e^{c(\textbf{g}^{\mathcal{V}}_j,\textbf{g}^{\mathcal{T}}_{i}) /\tau}}
    \right),
    \label{eq:infonce_loss}
\end{equation}
where the $\tau$ is temperature term. 
Additionally, we employ an identity classification loss $\mathcal{L}_\textrm{ID}$ to ensure that the embeddings extracted from the same identity become similar.  
The $\mathcal{L}_\textrm{ID}$ is denoted by
\begin{eqnarray}
\mathcal{L}_\textrm{ID} = -\sum_{i=1}^{B}\left( \boldsymbol{y}_i \log\sigma(\textbf{g}_i^{\mathcal{V}}\mathbf{W}_\textrm{ID}) 
 + \boldsymbol{y}_i \log\sigma(\textbf{g}_i^{\mathcal{T}}\mathbf{W}_\textrm{ID}) \right),
\label{eq:id loss}
\end{eqnarray}
where $\boldsymbol{y}_i\in \mathbb{R}^C$ is the identity ground truth of corresponding global embedding $\mathbf{g}_i$ represented by a one-hot vector and $C$ is the number of identities, $\mathbf{W}_\textrm{ID} \in \mathbb{R}^{D\times C}$ is a classifier shared between the two modalities.
Finally, we describe global alignment loss as below:
\begin{equation}
    \mathcal{L}_{\textrm{Global}} = \mathcal{L}_{\textrm{NCE}} + \mathcal{L}_{\textrm{ID}}.
    \label{eq:global_align_loss}
\end{equation}

\vspace{1mm}\noindent\textbf{Part Alignment Loss:}
Similar to the alignment of global embeddings, we first define a batch of part embeddings $\mathcal{B}_\text{part} = \{(\textbf{P}^{\mathcal{V}}_i, \textbf{P}^{\mathcal{T}}_i)\}_{i = 1}^B$ with batch size $B$, and we adopt the InfoNCE loss to align part embeddings extracted from two modalities.
For learning text query-based informative alignment of part embeddings between modalities, we employ text query-based similarity aggregation function $c_\textrm{agg}$ in Eq.~\ref{eq: part similarity} to compute InfoNCE loss.
We termed this InfoNCE with $c_\textrm{agg}$ as PartNCE loss, and it is formulated by
\begin{equation}
    \fontsize{9}{11}\selectfont
    \mathcal{L}_{\textrm{PartNCE}}= -\sum_{i=1}^{B}\left(\textrm{log}
    \frac{e^{c_\text{agg}(\textbf{P}^{\mathcal{V}}_i,\textbf{P}^{\mathcal{T}}_{i};~ \textbf{g}^{\mathcal{T}}_{i}) /\tau}}
        {\sum_{j=1}^B e^{c_\text{agg}(\textbf{P}^{\mathcal{V}}_i,\textbf{P}^{\mathcal{T}}_j;~ \textbf{g}^{\mathcal{T}}_j) /\tau}}
        + 
    \textrm{log}
    \frac{e^{c_\text{agg}(\textbf{P}^{\mathcal{V}}_i,\textbf{P}^{\mathcal{T}}_{i};~ \textbf{g}^{\mathcal{T}}_{i}) /\tau}}
        {\sum_{j=1}^B e^{c_\text{agg}(\textbf{P}^{\mathcal{V}}_j,\textbf{P}^{\mathcal{T}}_{i};~ \textbf{g}^{\mathcal{T}}_{i}) /\tau}}
    \right),
    \label{eq:partnce_loss}
\end{equation}
Similar to global alignment loss, we adopt the identity loss that shares classifier weights between the two modalities for part alignment; however, the difference is that the part embeddings are concatenated along the embedding dimension.
The identity loss for part alignment is formulated by
\begin{eqnarray}
\mathcal{L}_\textrm{PartID} = -\sum_{i=1}^{B}\left( \boldsymbol{y}_i \log\sigma([\textbf{P}_i^{\mathcal{V}}]\mathbf{W}_\textrm{PartID})
 + \boldsymbol{y}_i \log\sigma([\textbf{P}_i^{\mathcal{T}}]\mathbf{W}_\textrm{PartID})\right),
\label{eq:partid_loss}
\end{eqnarray}
where $[\mathbf{P}] \in \mathbb{R}^{KD}$ is the concatenation of part embeddings $\mathbf{P}$ along the embedding dimension, and $\mathbf{W}_\textrm{PartID} \in \mathbb{R}^{KD\times C}$ is a classifier shared between both modalities.
Finally, the part alignment loss is computed by
\begin{equation}
    \mathcal{L}_{\textrm{Part}} = \mathcal{L}_{\textrm{PartNCE}} + \mathcal{L}_{\textrm{PartID}}.
    \label{eq:part_align_loss}
\end{equation}

\begin{wrapfigure}{r}{0.5\textwidth}
\vspace{-12.5mm}
\begin{center}
\includegraphics[width = 0.5\textwidth]{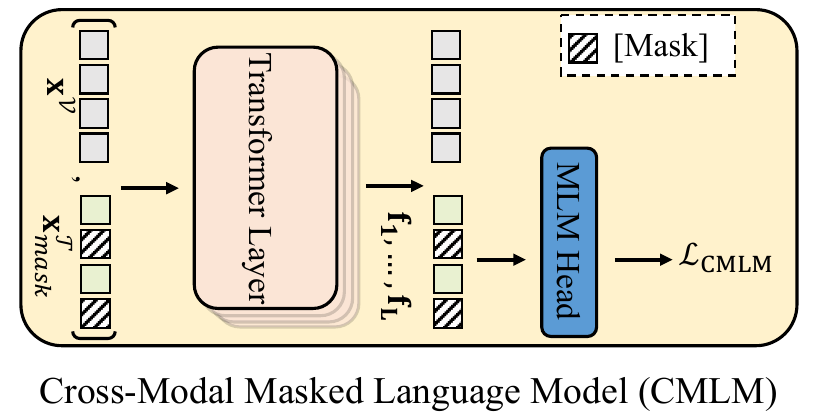}
\end{center}
\vspace{-6.5mm}
\caption{
Illustration of the CMLM.} 
\label{fig:CMLM}
\vspace{-9mm}
\end{wrapfigure}
\vspace{1mm}\noindent\textbf{Cross-Modal Masked Language Modeling Loss:}
Following the practices of employing Transformer-based backbones in existing methods~\cite{ivt, irra}, we adopt an auxiliary loss, the cross-modal masked language modeling (CMLM) loss, to facilitate the learning of interactions between modalities.
Similar to BERT~\cite{devlin2018bert}, given a text description, we randomly select text tokens with a 15\% probability and replace them with the learnable [\texttt{MASK}] token.
The masked text description is then processed through the text backbone to obtain the masked textual tokens.
After obtaining the masked textual tokens, we concatenate them with the visual tokens extracted via the image backbone.
The concatenated tokens are then fed into a transformer to acquire cross-modal fused tokens. 
Among the fused tokens, those corresponding to the indices of textual tokens are denoted as 
$\mathbf{F}$ composed as $\{ \mathbf{f}_1, \cdots \mathbf{f}_L \}$, where a fused token $\mathbf{f}\in \mathbb{R}^{1\times D}$ and $L$ indicates the max length of input text descriptions.
Ultimately, these tokens are fed into a CMLM classifier $\mathbf{W}_\textrm{CMLM}$ to predict the probability of vocabulary IDs.
The overall procedure of CMLM is illustrated in Fig.~\ref{fig:CMLM} and through this procedure we compute the $\mathcal{L}_\textrm{CMLM}$ loss as follows:
\begin{equation}
    \mathcal{L}_{\textrm{CMLM}} = -\frac{1}{L}\sum_{l=1}^{L}\boldsymbol{y}_l\log\big(\sigma(\mathbf{f}_{l}\mathbf{W}_{\textrm{CMLM}})\big),
    \label{eq:cmlm_loss}
\end{equation}
where $\mathbf{W}_\textrm{CMLM} \in \mathbb{R}^{D\times V}$ and $\boldsymbol{y}_l\in \mathbb{R}^V$ is the vocabulary ground truth of $l$-th text tokens, represented by a one-hot vector and $V$ is the size of vocabulary.
Finally, our overall objective function for training is denoted by
\begin{equation}
    \mathcal{L} = \mathcal{L}_{\textrm{Global}} + \mathcal{L}_{\textrm{Part}}  + \mathcal{L}_{\textrm{CMLM}}.
    \label{eq:overall_objective}
\end{equation}

\subsection{Inference}\label{sec:inference}
During testing, the global and part embeddings of each modality input are fully exploited to calculate the similarity between the image-text pair.
In particular, the similarities between visual part embeddings and their corresponding textual part embeddings are linearly combined with the attention weights $\textbf{\textrm{a}}$ to aggregate them. 
The image-text pair similarity is defined as the sum of the similarity between the global embeddings of the image-text pair and the similarity between the part embeddings of it, which can be computed by $c(\mathbf{g}^{\mathcal{V}},\mathbf{g}^{\mathcal{T}}) + c_{\textrm{agg}}(\mathbf{P}^{\mathcal{V}},\mathbf{P}^{\mathcal{T}};\mathbf{g}^{\mathcal{T}})$.
Finally, given the text query, the images in the gallery are ranked according to similarity scores between the images and the text for inference.

\section{Experiments}
\label{sec:Experiments}
In this section, we provide a detailed account of our experimental setup~(Sec.~\ref{subsec:experiment_setup}), evaluate our method, and compare it with state of the arts on three benchmark datasets for text-based person search~(Sec.~\ref{sec:quan}).
Furthermore, we qualitatively present retrieval results and analyze the effectiveness of the part discovery module and TDPA with visualization results (Sec.~\ref{sec:qual}). 
We also conduct ablation studies on the losses employed in model training, the methodologies of part discovery, and the strategies for part similarity aggregation (Sec.~\ref{sec:ablation}).

\subsection{Experimental Setup}
\label{subsec:experiment_setup}
\noindent \textbf{Datasets:} 
On three benchmark datasets, CUHK-PEDES \cite{li2017person}, ICFG-PEDES~\cite{ding2021semantically}, and RSTPReid~\cite{zhu2021dssl}, we evaluate and compare the performance of our method against previous methods.
In CUHK-PEDES collected from five existing person re-identification datasets~\cite{li2014deepreid, Market1501, xiao2016end, gray2007evaluating, li2013human}, it contains 40,206 images corresponding to 13,003 individual IDs, with each image being approximately matched with two annotated text descriptions.
We follow the data split of~\cite{li2017person} with 34,054 images from 11,003 person IDs and 68,126 text descriptions for training, 3,078 images from 1,000 IDs and 6,158 text descriptions for validation, and 3,074 images from 1,000 IDs and 6,156 text descriptions for testing.
The remaining two datasets are collected from MSMT17~\cite{wei2018person}.
ICFG-PEDES consists of 54,522 image-text pairs from 4,102 individual IDs, which are split into 34,674 and 19,848 for training and testing, respectively.
RSTPReid contains 20,505 images of 4,101 individual IDs, with each ID having 5 images and each image associated with the corresponding two annotated text descriptions.
We follow the data split of~\cite{zhu2021dssl} with 18,505 images from 3,701 IDs and 37,010 text descriptions for training, 1,000 images from 200 IDs and 2,000 text descriptions for validation, and 1,000 images from 200 IDs and 2,000 text descriptions for testing, respectively.

\noindent \textbf{Evaluation Protocol:}
We employ the standard metric of rank at K (R@K=1,5, 10) for all retrieval experiments.
Specifically, given a query text, images are sorted based on their similarity to the query text. The search is considered correct if at least one relevant image appears in the top K positions of the ranking.

\noindent{\textbf{Network Architecture:}} 
We adopt the pre-trained CLIP models from OpenAI~\cite{clip} for both image and text encoders, where the size of the image encoder is ViT-B/16.
The input images are resized to 384$\times$128.
Random horizontal flipping, random cropping, and random erasing are applied for the data augmentation in training time.
The maximum text length is set to 77.

\noindent{\textbf{Network Optimization:}} 
Our model is trained using the Adam optimizer for 60 epochs with a batch size of 128 for all experiments.
For the CLIP encoders, the initial learning rate is set to $\expnum{5}{6}$, using a cosine schedule with the warm-up strategy at the first five epochs; we use a high learning rate for the remaining parameters by scaling 20 times.

\noindent{\textbf{Hyperparameters:}} We set the number of part slots to 8. The number of iterations in the part slot attention block is set to 5.
The temperature parameter $\tau$ is set to 0.015. 


\begin{table*}[!t]
    \centering
    \caption{Performance of text-based person search methods on the three datasets. \textbf{Bold} and \underline{underline} denote the best and the second best.} 
    \vspace{-3mm}
    \fontsize{7.5}{9.5}\selectfont
    \resizebox{1.0\textwidth}{!}{%
    \begin{tabular}{l|cc|ccc|ccc|ccc}
    \toprule
    &\multicolumn{2}{c|}{\textbf{Backbone}}&\multicolumn{3}{c|}{\textbf{CUHK-PEDES}} &\multicolumn{3}{c|}{\textbf{ICFG-PEDES}}& \multicolumn{3}{c}{\textbf{RSTPReid}} \\[-0.3ex]  \cmidrule(lr){2-3} \cmidrule(lr){4-6} \cmidrule(lr){7-9} \cmidrule(lr){10-12}
    \textbf{Methods} &\multicolumn{1}{c}{Image} &\multicolumn{1}{c|}{Text} & \multicolumn{1}{c}{R@1} & \multicolumn{1}{c}{R@5} & \multicolumn{1}{c|}{R@10} & \multicolumn{1}{c}{R@1} & \multicolumn{1}{c}{R@5} & R@10 & \multicolumn{1}{c}{R@1} & \multicolumn{1}{c}{R@5} & \multicolumn{1}{c}{R@10} \\ [-0.3ex] \midrule
    GNA-RNN~\cite{li2017person} & RN50 & LSTM & 19.05 & - & 53.64 &-&-&-& -& -& -\\
    CMPM/C~\cite{zhang2018deep} & RN50 & LSTM  & 49.37 & 71.69 & 79.27 & 43.51& 65.44& 74.26&- &- &- \\
    PMA~\cite{jing2020pose}& RN50 & BERT  & 53.81 & 73.54 & 81.23 & - & - & - & -&- &- \\
    TIMAM~\cite{sarafianos2019adversarial} & RN101 & BERT  & 54.51 & 77.56 & 84.78 &- & -& -& -& -& -\\
    SCAN~\cite{lee2018stacked}& RN50 & BERT & 55.86 & 75.97 & 83.69 & 50.05& 69.65& 77.21& -&- &- \\
    ViTAA~\cite{wang2020vitaa}& RN50 & LSTM & 55.97 & 75.84 & 83.52 & 50.98& 68.79& 75.78& -& -&- \\
    NAFS~\cite{gao2021contextual}& RN50 & BERT & 59.94 & 79.86 & 86.70 & - & - & - & -&- &- \\
    DSSL~\cite{zhu2021dssl}& RN50 & BERT & 59.98 & 80.41 &87.56& - & - & - & 32.43 & 55.08 & 63.19 \\
    MGEL~\cite{wang2021text}& RN50 & LSTM & 60.27 & 80.01 & 86.74 & - & - & - & -&- &- \\
    SSAN~\cite{ding2021semantically}& RN50 & LSTM & 61.37 & 80.15 & 86.73 & 54.23 & 72.63 & 79.53& 43.50 & 67.80 & 77.15 \\
    LapsCore~\cite{wu2021lapscore}& RN50 & BERT & 63.40 & - & 87.80 & - & - & - & -&- &- \\
    SRCF~\cite{suo2022simple} & RN50 & BERT & 64.04 & 82.99 & 88.81 & 57.18&  {75.01}&  {81.49}& - &  -&  -\\
    LGUR~\cite{lgur} & RN50 & BERT & 64.21 & 81.94 & 87.93 &  {57.42} & 74.97 & 81.45 & -&- &- \\
    TIPCB~\cite{chen2022tipcb}& RN50 & BERT & 64.26 &  {83.19} &  {89.10} & - & - & - & -&- &- \\
    CAIBC~\cite{caibc}& RN50 & BERT &  {64.43} & 82.87 & 88.37 & - & - & - &  {47.35}& 69.55 & 79.00 \\
    SAF~\cite{saf} & ViT-B/16 & BERT & 64.13 &82.62 &88.40 & - & - & - & - & - & -\\
    IVT~\cite{ivt} & ViT-B/16 & BERT &  {65.59} & {83.11} & {89.21} &  {56.04} & {73.60} & {80.22} &  {46.70} & {70.00} &{78.80}\\
    CFine~\cite{CFine} & CLIP-ViT-B/16 & BERT &  69.57 & 85.93 & 91.15 &  60.83 & 75.55 & 82.42 &  50.55 & 72.50 & 81.60\\
    IRRA~\cite{irra} & CLIP-ViT-B/16 & CLIP-Xformer &  {73.38} & {89.93} & {93.71} &  {63.46} & {80.24} & {85.82} &  {60.20} & {	81.30} &{88.20}\\ 
    \ccol Ours & \ccol CLIP-ViT-B/16 &\ccol  CLIP-Xformer &\ccol \textbf{75.28} &\ccol \textbf{90.42} &\ccol \textbf{94.12} &\ccol \textbf{65.76} &\ccol \textbf{81.39}  &\ccol \textbf{86.73} &\ccol \textbf{61.80}&\ccol \textbf{82.85}&\ccol \textbf{89.45} \\ \bottomrule
    \end{tabular}
    }
    \label{tab:quan}
    \vspace{-6mm}
\end{table*}

\subsection{Quantitative Results}
\label{sec:quan}
We compare our method with previous text-based person search methods on CUHK-PEDES~\cite{li2017person}, ICFG-PEDES~\cite{ding2021semantically}, and RSTPReid~\cite{zhu2021dssl}.
The performance comparison and the backbones for each modality employed by each method are shown in Table~\ref{tab:quan}.
Specifically, our method achieves an outstanding R@1 metric of 75.28\%, 65.76\%, and 61.80\% on the CUHK-PEDES, ICFG-PEDES, and RSTPReid datasets, respectively, outperforming previous methods utilizing different backbones from ours.
Moreover, our method improves the previous state of the art, IRRA~\cite{irra}, on R@1 by a large margin as 1.9\%p, 2.3\%p, and 1.6\%p, respectively. 
Since IRRA only focuses on aligning the global embeddings of each modality, it is hard to capture the fine-grained differences.
In contrast, our method not only takes account of the global embeddings but also aims to discover and align discriminative part embeddings from each modality; it allows the model to effectively find the target person.

\subsection{Qualitative Results}
\label{sec:qual}
\begin{figure*}[t!]
    \centering
    \includegraphics[width=0.92\textwidth]{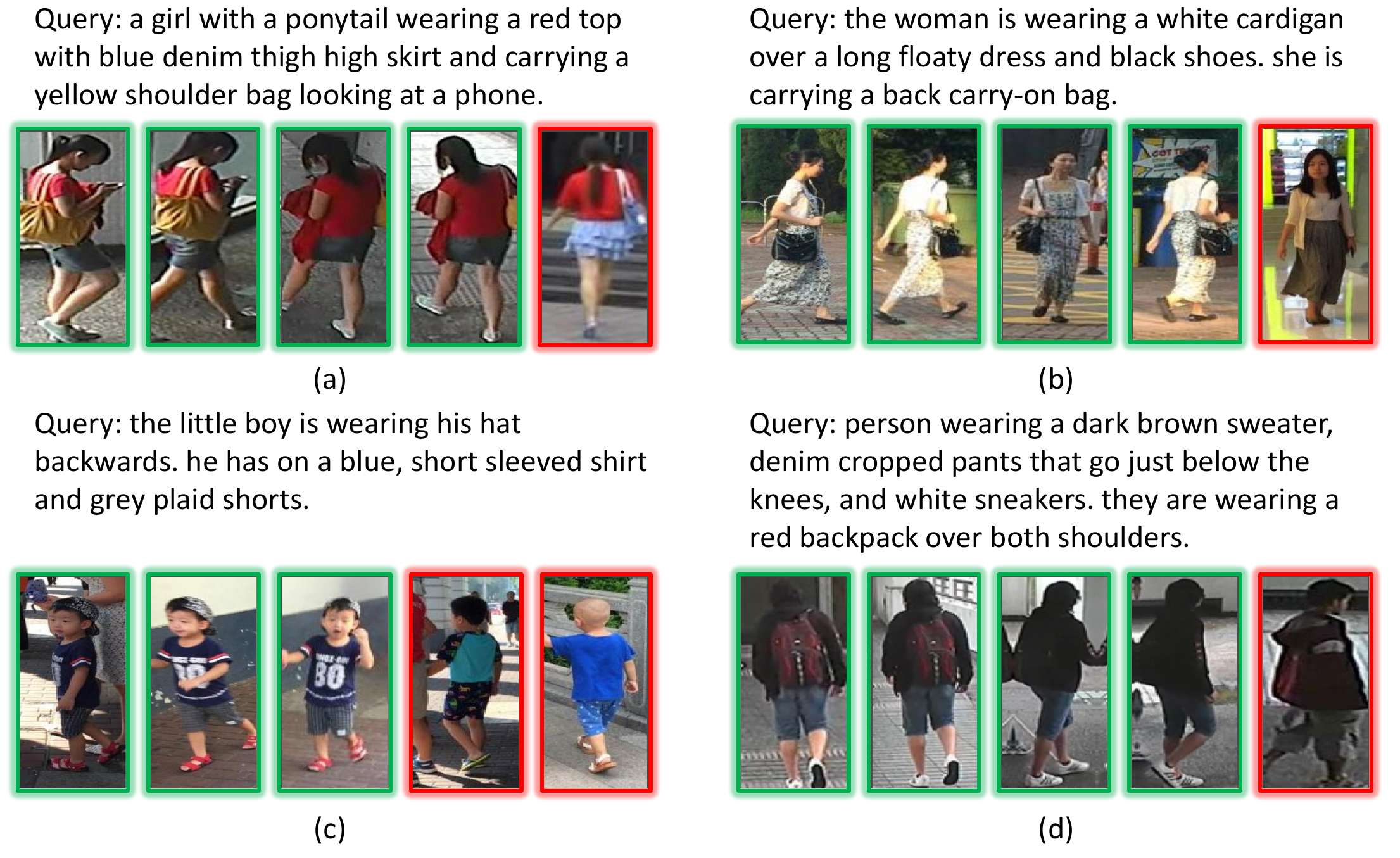}
    \vspace{-3mm}
    \caption{
    Top-5 retrieval results of our method on the CUHK-PEDES dataset.
    Images are sorted from left to right according to their ranks below each text query.
    Green and red boxes indicate true and false matches, respectively.
    }
    \vspace{-7mm}
    \label{fig:qualitative}
\end{figure*}

\noindent{\textbf{Retrieval Results:}}
Top-5 retrieval results of our method on the CUHK-PEDES dataset are illustrated in Fig.~\ref{fig:qualitative}.
Above all, it shows the overall satisfactory retrieval results.
In particular, we can observe our model retrieves targets well, even with distinctive human parts that are small or located in various positions.
For instance, the small distinctive human parts like ``\sftype{ponytails}'' and ``\sftype{high skirts}'' in Fig.~\ref{fig:qualitative}(a), ``\sftype{black shoes}'' in Fig.~\ref{fig:qualitative}(b), and ``\sftype{blue plaid shorts}'' in Fig.~\ref{fig:qualitative}(c), as well as human parts that could appear in various viewpoints such as ``\sftype{yellow shoulder bags}'' in Fig.~\ref{fig:qualitative}(a), ``\sftype{floaty dresses}'' in Fig.~\ref{fig:qualitative}(b), and ``\sftype{red backpacks}" in Fig.~\ref{fig:qualitative}(d).
The CUHK-PEDES dataset typically contains three target images on average in the search space, thereby most of the false matches in the figure are observed by a lack of additional targets.
Despite the false matching due to the limitation of the dataset, the retrieval results are reasonable in that the retrieved false matching contains distinctive human parts described by the query description.

\begin{figure*}[t!]
    \centering
    \includegraphics[width=0.95\textwidth]{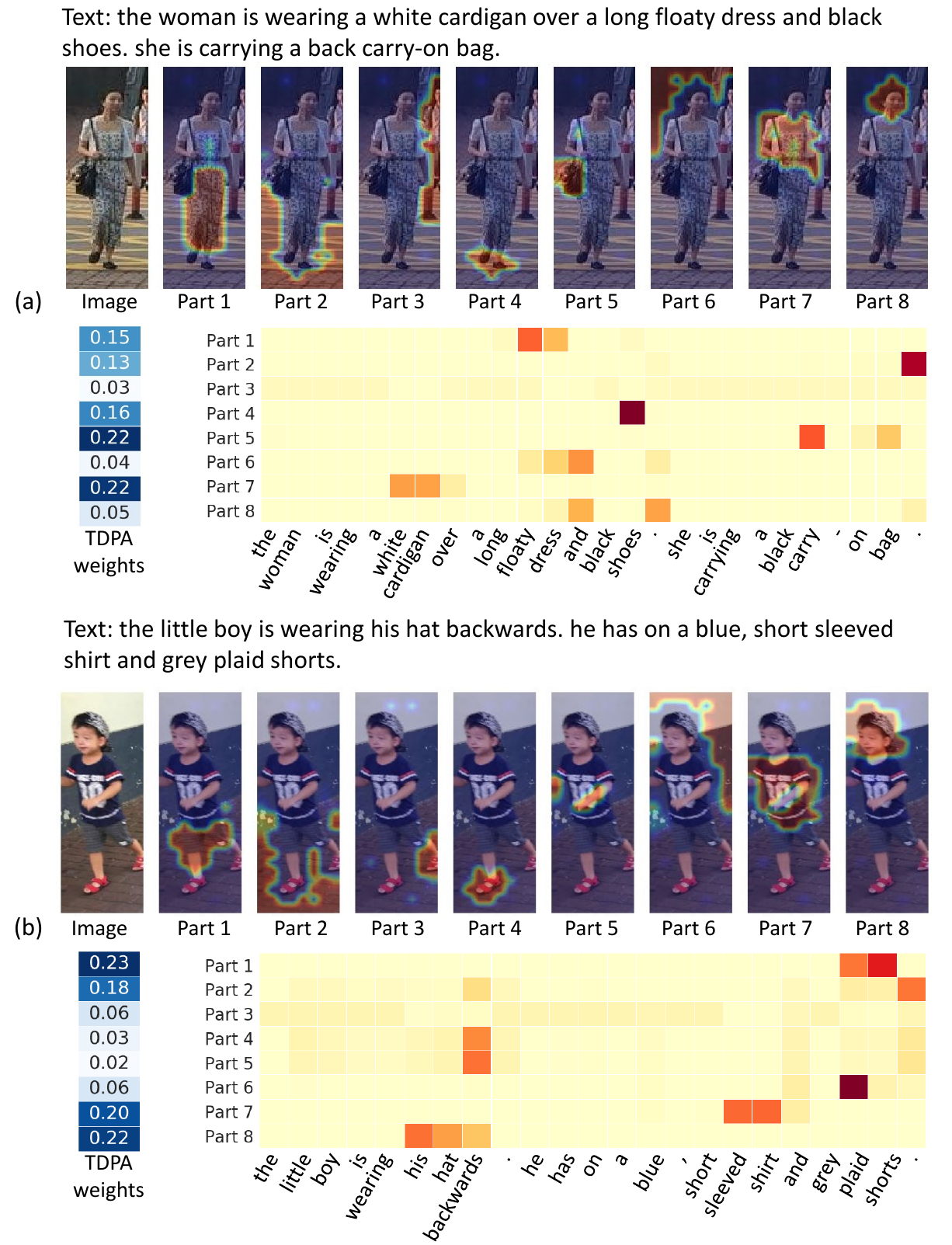}
    \vspace{-2mm}
    \caption{Visualization of each modality's attention map $\bar{A}_{k}$ in $T$-th iteration of PSA block and TDPA weights $\mathbf{a}$ on CUHK-PEDES dataset.
    }
    \vspace{-8.5mm}
    \label{fig:vis_attn}
\end{figure*}

\noindent{\textbf{Visualization of Attention Map $\bar{A}_{k}$ in PSABlock:}}
To demonstrate the effectiveness of our part discovery module, we visualize the attention map $\bar{A}_{k}$ in $T$-th iteration of the PSA block for both visual and textual modalities (in Eq.~\eqref{eq:part_slot_attention}).
The visualization results are illustrated in Fig.~\ref{fig:vis_attn}.
It not only demonstrates that the part embeddings extracted by our part discovery module capture distinctive human parts but also shows that the part embeddings extracted from the same part slot attend to the semantically identical human parts regardless of modality.
For example, the 1st part slot typically focuses on bottom clothes, the 4th on footwear, the 5th on objects being held, the 7th on top clothes, and the 8th on the person's head.
Furthermore, in Fig.~\ref{fig:vis_attn} compared (a) and (b) of visual modality, our part discovery module is capable of capturing the distinctive human parts, while it is robust against pose variations and viewpoint changes.

\noindent{\textbf{Visualization of TDPA:}}
We visualize TDPA weights $\mathbf{a}$ in Eq.~\eqref{eq: part similarity} to demonstrate the effectiveness of similarity aggregation between part embeddings of two modalities with TDPA.
The visualization results are presented in the bottom left in Fig.~\ref{fig:vis_attn}(a, b), respectively.
Comparing examples (a) and (b), the 5th part slot typically focuses on human parts associated with held objects, like bags, leading to a predicted lower TDPA weight for this part slot in (b) due to the absence of such distinctive information in its query description.
However, the presence of the ``\sftype{hat}" in the query description of (b) leads to having a high TDPA weight for the 8th part slot that typically focuses on the human head part.
In contrast, since this related human head part information is not provided in the query description of (a), the TDPA weight for the 8th part slot in (a) is predicted to be a low value.
These observations highlight the capability of TDPA, which adaptively improves part-based retrieval depending on the contents of the text query.

\subsection{Ablation Studies}
\label{sec:ablation}
In our ablation studies conducted on the CUHK-PEDES datasets, we evaluate the effectiveness of proposed components and their combinations in improving text-based person search performance.
In Table~\ref{tab:ablation_component}, we compare several configurations: The baseline method employs only global embeddings $\mathbf{g}$ trained by InfoNCE ($\mathcal{L}_{\textrm{NCE}}$) loss.
When incorporating the cross-modal masked language modeling ($\mathcal{L}_{\textrm{CMLM}}$), there is a slight improvement across all metrics.
The addition of part embeddings $\mathbf{P}$ that without $\mathcal{L}_{\textrm{PartID}}$ also shows significant enhancement in R@1 and R@5, where we observe an increase of 3.46\%p and 2.64\%p, respectively.
Notably, the full configuration achieves the best performance with $\mathcal{L}_{\textrm{PartID}}$, underscoring the importance of part embeddings for accurate text-based person search.
This suggests that our part embeddings significantly contribute to discerning fine-grained details critical for text-based person search.
Additionally, the effectiveness of TDPA, along with other ablation studies, is presented in Table~\ref{tab:more_ablation} of our supplementary materials for a more comprehensive analysis.
\begin{table*}[!t]
    \centering
\caption{Ablation studies on the CUHK-PEDES datasets.}
    \vspace{-3.5mm}
    \resizebox{0.95\textwidth}{!}{   
    \begin{tabular}{lcccccccc}
    \toprule
     & \multicolumn{5}{c}{\textbf{Loss}} & \multicolumn{3}{c}{\textbf{CUHK-PEDES}}  \\[-0.3ex]  \cmidrule(lr){2-6} \cmidrule(lr){7-9} 
     \textbf{Method}& $\mathcal{L}_{\textrm{NCE}}$   & $\mathcal{L}_{\textrm{ID}}$ &$\mathcal{L}_{\textrm{CMLM}}$  &$\mathcal{L}_{\textrm{PartNCE}}$&$\mathcal{L}_{\textrm{PartID}}$ &\multicolumn{1}{c}{R@1} & \multicolumn{1}{c}{R@5} & \multicolumn{1}{c}{R@10}  \\ [-0.3ex] \midrule
    \multirow{2}{*}{Global Only} & \cmark & \xmark & \xmark &\xmark&\xmark&71.39 (+0.0)& 87.65 (+0.0)& 92.74 (+0.0) \\
    & \cmark &\cmark &\xmark &\xmark&\xmark&71.83 \textcolor{grn}{(+0.44)}& 88.06 \textcolor{grn}{(+0.41)}& 92.58 \textcolor{red}{(-0.16)} \\ \midrule
    $+$ CMLM& \cmark &\cmark &\cmark&\xmark&\xmark&72.65 \textcolor{grn}{(+1.26)}& 88.58 \textcolor{grn}{(+0.93)}& 92.93 \textcolor{grn}{(+0.19)}  \\ \midrule
    \multirow{2}{*}{$+$ Part Embeddings}& \cmark &\cmark &\cmark &\cmark&\xmark& 74.85 \textcolor{grn}{(+3.46)}  & 90.29 \textcolor{grn}{(+2.64)}  & 94.10 \textcolor{grn}{(+1.36)}   \\
     & \cmark  & \cmark &\cmark &\cmark&\cmark& \textbf{75.28} \textcolor{grn}{(+3.89)} & \textbf{90.42} \textcolor{grn}{(+2.77)} & \textbf{94.12} \textcolor{grn}{(+1.38)} \\ \bottomrule
    \end{tabular}
    }
    \vspace{-6mm}
    \label{tab:ablation_component}
\end{table*}


\begin{wraptable}{r}{5.8cm}
    \vspace{-11.5mm}
    \caption{Ablation study of different part discovery methods on the CUHK-PEDES.}
    \label{tab:ablation_similarity_aggregation}
    \resizebox{0.49\textwidth}{!}{%
    \begin{tabular}{clccc}
    \toprule
    \multicolumn{1}{c}{}&\multicolumn{1}{l}{\textbf{Methods}}  & R@1 & R@5 & R@10\\ [-0.3ex]  \midrule
    \multicolumn{1}{c}{}&\multicolumn{1}{l}{Ours + TIPCB~\cite{chen2022tipcb}} & 73.23 &  89.10 & 94.04 \\
    \multicolumn{1}{c}{}&\multicolumn{1}{l}{Ours + PAT~\cite{pat}} & 72.76 &  89.23 & 93.42 \\ \midrule
    \multicolumn{1}{c}{}&\multicolumn{1}{l}{Ours}& \textbf{75.28} & \textbf{90.42} & \textbf{94.12} \\ \bottomrule
    \end{tabular}
    }
\vspace{-7.5mm}
\end{wraptable}

\noindent{\textbf{Ablation Study on Part Discovery Methods:}}
To validate the effectiveness of our proposed part discovery module, we conduct experiments by replacing our part discovery module with other methods such as TIPCB~\cite{chen2022tipcb} and PAT~\cite{pat}.
TIPCB is a simple yet effective heuristic part discovery method, which extracts image part embeddings through equi-horizontal cropping of the entire image and transforms the global embedding of the text modality into corresponding parts by using additional learnable MLPs for each image part.
PAT performs part discovery with a querying transformer approach by leveraging learnable queries and conventional cross-attention.
A key difference of our method from PAT is that ours explicitly encourages the discovered parts to be spatially separated since our part slots compete with each other to aggregate input data.
In contrast, parts found by PAT, based on conventional cross-attention, often over-capture salient regions and are likely to miss fine details.
To demonstrate this limitation, we show the visualization of part attentions with PAT applied to ours in Fig.~\ref{fig:supp_vis_pat} of the supplementary materials.
As indicated in Table~\ref{tab:ablation_similarity_aggregation}, our part discovery method outperforms these heuristic and conventional cross-attention based part discovery methods across all metrics, with the most significant difference observed in the R@1 metric, which requires a precise discernment of human parts.

\section{Conclusion}
We proposed a novel framework that extracts distinctive human parts corresponding across visual and textual modalities through part discovery module without part-level supervision.
The introduced TDPA further refines the retrieval process by adjusting the importance of each part slot based on the contents of the text query, leading to more precise and relevant retrieval.

\noindent \textbf{Limitation:} Since slots learn to occupy the entire image and text, some of them may indicate irrelevant part of input. TDPA addresses this issue to some extent, but a more explicit solution would further enhance our system.
\label{sec:conclusion}

\bigskip
{\small
\noindent \textbf{Acknowledgement:}
This work was partly supported by the IITP grants and the NRF grants funded by Ministry
of Science and ICT, Korea (No.RS-2019-II191906 AI Graduate School Program (POSTECH); No.RS-2022-II220906
}

\bibliographystyle{splncs04}
\bibliography{cvlab_kwak}

\begin{thebibliography}{10}
\providecommand{\url}[1]{\texttt{#1}}
\providecommand{\urlprefix}{URL }
\providecommand{\doi}[1]{https://doi.org/#1}

\bibitem{aggarwal2020text}
Aggarwal, S., Radhakrishnan, V.B., Chakraborty, A.: Text-based person search via attribute-aided matching. In: Proc. Winter Conference on Applications of Computer Vision (WACV) (2020)

\bibitem{bai2023rasa}
Bai, Y., Cao, M., Gao, D., Cao, Z., Chen, C., Fan, Z., Nie, L., Zhang, M.: Rasa: Relation and sensitivity aware representation learning for text-based person search. In: Proc. International Joint Conferences on Artificial Intelligence (IJCAI) (2023)

\bibitem{chen2022tipcb}
Chen, Y., Zhang, G., Lu, Y., Wang, Z., Zheng, Y.: Tipcb: A simple but effective part-based convolutional baseline for text-based person search. Neurocomputing  \textbf{494} (2022)

\bibitem{devlin2018bert}
Devlin, J., Chang, M.W., Lee, K., Toutanova, K.: Bert: Pre-training of deep bidirectional transformers for language understanding. In: Annual Conference ofthe North American Chapter ofthe Association for Computational Linguistics (2019)

\bibitem{ding2021semantically}
Ding, Z., Ding, C., Shao, Z., Tao, D.: Semantically self-aligned network for text-to-image part-aware person re-identification. arXiv preprint arXiv:2107.12666  (2021)

\bibitem{gao2021contextual}
Gao, C., Cai, G., Jiang, X., Zheng, F., Zhang, J., Gong, Y., Peng, P., Guo, X., Sun, X.: Contextual non-local alignment over full-scale representation for text-based person search. arXiv preprint arXiv:2101.03036  (2021)

\bibitem{gray2007evaluating}
Gray, D., Brennan, S., Tao, H.: Evaluating appearance models for recognition, reacquisition, and tracking. In: Proc. IEEE international workshop on performance evaluation for tracking and surveillance (PETS) (2007)

\bibitem{irra}
Jiang, D., Ye, M.: Cross-modal implicit relation reasoning and aligning for text-to-image person retrieval. In: IEEE International Conference on Computer Vision and Pattern Recognition (CVPR) (2023)

\bibitem{jing2020pose}
Jing, Y., Si, C., Wang, J., Wang, W., Wang, L., Tan, T.: Pose-guided multi-granularity attention network for text-based person search. In: Proc. AAAI Conference on Artificial Intelligence (AAAI) (2020)

\bibitem{lee2018stacked}
Lee, K.H., Chen, X., Hua, G., Hu, H., He, X.: Stacked cross attention for image-text matching. In: Proc. European Conference on Computer Vision (ECCV) (2018)

\bibitem{ALBEF}
Li, J., Selvaraju, R.R., Gotmare, A.D., Joty, S., Xiong, C., Hoi, S.: Align before fuse: Vision and language representation learning with momentum distillation. In: Proc. Neural Information Processing Systems (NeurIPS) (2021)

\bibitem{saf}
Li, S., Cao, M., Zhang, M.: Learning semantic-aligned feature representation for text-based person search. In: Proc. IEEE International Conference on Acoustics, Speech and Signal Processing (ICASSP). IEEE (2022)

\bibitem{li2017person}
Li, S., Xiao, T., Li, H., Zhou, B., Yue, D., Wang, X.: Person search with natural language description. In: Proc. IEEE Conference on Computer Vision and Pattern Recognition (CVPR) (2017)

\bibitem{li2013human}
Li, W., Zhao, R., Wang, X.: Human reidentification with transferred metric learning. In: Computer Vision--ACCV 2012: 11th Asian Conference on Computer Vision, Daejeon, Korea, November 5-9, 2012, Revised Selected Papers, Part I 11. Springer (2013)

\bibitem{li2014deepreid}
Li, W., Zhao, R., Xiao, T., Wang, X.: Deepreid: Deep filter pairing neural network for person re-identification. In: Proc. IEEE Conference on Computer Vision and Pattern Recognition (CVPR) (2014)

\bibitem{pat}
Li, Y., He, J., Zhang, T., Liu, X., Zhang, Y., Wu, F.: Diverse part discovery: Occluded person re-identification with part-aware transformer. In: Proc. IEEE Conference on Computer Vision and Pattern Recognition (CVPR) (2021)

\bibitem{slot}
Locatello, F., Weissenborn, D., Unterthiner, T., Mahendran, A., Heigold, G., Uszkoreit, J., Dosovitskiy, A., Kipf, T.: Object-centric learning with slot attention. In: Proc. Neural Information Processing Systems (NeurIPS) (2020)

\bibitem{niu2020improving}
Niu, K., Huang, Y., Ouyang, W., Wang, L.: Improving description-based person re-identification by multi-granularity image-text alignments. IEEE Transactions on Image Processing  \textbf{29},  5542--5556 (2020)

\bibitem{infonce}
Oord, A.v.d., Li, Y., Vinyals, O.: Representation learning with contrastive predictive coding. arXiv preprint arXiv:1807.03748  (2018)

\bibitem{clip}
Radford, A., Kim, J.W., Hallacy, C., Ramesh, A., Goh, G., Agarwal, S., Sastry, G., Askell, A., Mishkin, P., Clark, J., et~al.: Learning transferable visual models from natural language supervision. In: Proc. International Conference on Machine Learning (ICML) (2021)

\bibitem{sarafianos2019adversarial}
Sarafianos, N., Xu, X., Kakadiaris, I.A.: Adversarial representation learning for text-to-image matching. In: Proc. IEEE International Conference on Computer Vision (ICCV) (2019)

\bibitem{lgur}
Shao, Z., Zhang, X., Fang, M., Lin, Z., Wang, J., Ding, C.: Learning granularity-unified representations for text-to-image person re-identification. In: Proc. ACM Multimedia Conference (ACMMM) (2022)

\bibitem{ivt}
Shu, X., Wen, W., Wu, H., Chen, K., Song, Y., Qiao, R., Ren, B., Wang, X.: See finer, see more: Implicit modality alignment for text-based person retrieval. In: Proc. European Conference on Computer Vision Workshop on Real-World Surveillance, (ECCVW) (2022)

\bibitem{suo2022simple}
Suo, W., Sun, M., Niu, K., Gao, Y., Wang, P., Zhang, Y., Wu, Q.: A simple and robust correlation filtering method for text-based person search. In: Proc. European Conference on Computer Vision (ECCV) (2022)

\bibitem{wang2021text}
Wang, C., Luo, Z., Lin, Y., Li, S.: Text-based person search via multi-granularity embedding learning. In: Proc. International Joint Conferences on Artificial Intelligence (IJCAI) (2021)

\bibitem{wang2020vitaa}
Wang, Z., Fang, Z., Wang, J., Yang, Y.: Vitaa: Visual-textual attributes alignment in person search by natural language. In: Proc. European Conference on Computer Vision (ECCV) (2020)

\bibitem{caibc}
Wang, Z., Zhu, A., Xue, J., Wan, X., Liu, C., Wang, T., Li, Y.: Caibc: Capturing all-round information beyond color for text-based person retrieval. In: Proc. ACM Multimedia Conference (ACMMM) (2022)

\bibitem{watters2019spatial}
Watters, N., Matthey, L., Burgess, C.P., Lerchner, A.: Spatial broadcast decoder: A simple architecture for learning disentangled representations in vaes. In: ICLR Workshop (2019)

\bibitem{wei2018person}
Wei, L., Zhang, S., Gao, W., Tian, Q.: Person transfer gan to bridge domain gap for person re-identification. In: Proc. IEEE Conference on Computer Vision and Pattern Recognition (CVPR) (2018)

\bibitem{wu2021lapscore}
Wu, Y., Yan, Z., Han, X., Li, G., Zou, C., Cui, S.: Lapscore: language-guided person search via color reasoning. In: Proc. IEEE International Conference on Computer Vision (ICCV) (2021)

\bibitem{xiao2016end}
Xiao, T., Li, S., Wang, B., Lin, L., Wang, X.: End-to-end deep learning for person search. arXiv preprint arXiv:1604.01850  (2016)

\bibitem{CFine}
Yan, S., Dong, N., Zhang, L., Tang, J.: Clip-driven fine-grained text-image person re-identification. IEEE Transactions on Image Processing  (2023)

\bibitem{zhang2018deep}
Zhang, Y., Lu, H.: Deep cross-modal projection learning for image-text matching. In: Proc. European Conference on Computer Vision (ECCV) (2018)

\bibitem{Market1501}
Zheng, L., Shen, L., Tian, L., Wang, S., Wang, J., Tian, Q.: Scalable person re-identification: A benchmark. In: Proc. IEEE International Conference on Computer Vision (ICCV) (2015)

\bibitem{zhu2021dssl}
Zhu, A., Wang, Z., Li, Y., Wan, X., Jin, J., Wang, T., Hu, F., Hua, G.: Dssl: deep surroundings-person separation learning for text-based person retrieval. In: Proceedings of the 29th ACM International Conference on Multimedia. pp. 209--217 (2021)

\end{thebibliography}

\clearpage
\newcounter{alphasect}
\def\alphainsection{0}

\let\oldsection=\section
\def\section{%
  \ifnum\alphainsection=1%
    \addtocounter{alphasect}{1}
  \fi%
\oldsection}%

\renewcommand\thesection{%
  \ifnum\alphainsection=1%
    \Alph{alphasect}
  \else%
    \arabic{section}
  \fi%
}%

\newenvironment{alphasection}{%
  \ifnum\alphainsection=1%
    \errhelp={Let other blocks end at the beginning of the next block.}
    \errmessage{Nested Alpha section not allowed}
  \fi%
  \setcounter{alphasect}{0}
  \def\alphainsection{1}
}{%
  \setcounter{alphasect}{0}
  \def\alphainsection{0}
}%

\title{\paperTitleSupp}

\titlerunning{\abbvPaperTitle}
\author{}
\institute{}
\maketitle
\begin{alphasection}
This supplementary material presents auxiliary feature reconstruction loss function and experimental results omitted from the main paper due to the space limit.
We first describe the auxiliary feature reconstruction loss for the part discovery module in Sec.~\ref{sec:feat_recon}.
In Sec.~\ref{sec:comp_rasa}, we provide a comparison with Rasa~\cite{bai2023rasa}, a reranking-based text-based person search method that utilizes a classifier for calculating similarity scores between text queries and images.
Thereafter, Sec.~\ref{sec:more_ablation} offers more ablation studies on variants of our method, such as other similarity aggregation methods, the use of the GRU update in the PSA block, and varying the number of part slots.
Lastly, Sec.~\ref{sec:more_qual} presents additional qualitative results. Also, we pledge to release all associated code for our method and experimental evaluation to the public domain.

\vspace{0mm}\section{Feature Reconstruction for Part Discovery}
\label{sec:feat_recon}
To train our part discovery module, inspired by the object-centric learning method~\cite{slot}, we utilize an auxiliary loss for reconstruction.
When a set of part slots is able to reconstruct the original input data, it suggests that each slot is associated with unique part entities in the input.
For efficient reconstruction, we opt to conduct reconstruction in the feature space rather than directly reconstructing the raw input data; in other words, we reconstruct the outputs of the backbones for each input modality, namely the patch tokens $\mathbf{x}^{\mathcal{V}}$ and text tokens $\mathbf{x}^{\mathcal{T}}$.
To do this, we leverage the reconstruction loss $\mathcal{L}_{\textrm{recon}}$, which is formulated as follows:
\begin{equation*}
    \mathcal{L}_{\textrm{Recon}}= \frac{1}{B}\sum_{i=1}^{B}\big(\parallel \mathbf{x}_i^{\mathcal{V}}- f_\textrm{dec}^\mathcal{V}(\textbf{P}_i^{\mathcal{V}})  \parallel^2 + 
    \parallel \mathbf{x}_i^{\mathcal{T}}- f_\textrm{dec}^\mathcal{T}(\textbf{P}_i^{\mathcal{T}})  \parallel^2 \big),
    \label{eq:recon_loss}
\end{equation*}
where $\textbf{P}$ is represents the part embedding, output from the part discovery module for each modality, and  $f_\textrm{dec}$ is the decoder used for reconstruction in each modality.
The decoder $f_\textrm{dec}$ follows the structure of the spatial broadcast decoder~\cite{watters2019spatial} used in slot attention~\cite{slot}, which takes part embeddings as input and duplicates each part embedding a number of times corresponding to the target tokens for reconstruction.
Then, the duplicated part embeddings are added with learnable positional embeddings.
Finally, these enhanced part embeddings are fed into multi-layer percetron (MLP) with ReLU activation function to obtain aggregated reconstruction features across part embeddings by predicting target tokens and aggregation weights for each part embeddings on the duplicated positions.


\renewcommand{\thetable}{\Alph{table}}
\vspace{0mm}\section{Comparison to Recent Work}
\label{sec:comp_rasa}
RaSa~\cite{bai2023rasa} is one of the most recent text-based person search methods, which adopts the pre-trained ALBEF~\cite{ALBEF} model as its backbone.
In inference, RaSa initially computes cosine similarity using global embeddings between query texts and images, then re-ranks the top-k images per query.
The re-ranking process calculates cross-attention between each query text and its corresponding top-k images by employing an additional transformer-based model.
Then, the output of the transformer, a cross-attended feature obtained with cross attention, is fed into a classifier to predict a matching score of query and image pair.
Although the re-ranking process enhances retrieval accuracy, the necessity of an additional transformer feedforward pass adds a significant computational load during inference.
In Table~\ref{tab:comp_rasa}, we compare the proposed method and RaSa, in terms of Recall@$K$ and the latency of inference.
The results demonstrates that our method is at least 40 times faster during inference time, while maintaining or surpassing performance of RaSa in Recall@K metrics.

\begin{table*}[!t]
    \centering
    \caption{Comparison to RaSa~\cite{bai2023rasa} on the CUHK-PEDES and ICFG-PEDES datasets.} 
    \vspace{0mm}
    \fontsize{7.5}{9.5}\selectfont
    \resizebox{1.0\textwidth}{!}{
    \begin{tabular}{l|cccc|cccc|cccc}
    \toprule
    &\multicolumn{4}{c|}{\textbf{CUHK-PEDES}} &\multicolumn{4}{c|}{\textbf{ICFG-PEDES}} &\multicolumn{4}{c}{\textbf{RSTPReid}} \\[-0.3ex]  \cmidrule(lr){2-5} \cmidrule(lr){6-9} \cmidrule(lr){10-13}
    \textbf{Methods}  & \multicolumn{1}{c}{R@1} & \multicolumn{1}{c}{R@5} & \multicolumn{1}{c}{R@10} &\multicolumn{1}{c|}{Time(s)} & \multicolumn{1}{c}{R@1} & \multicolumn{1}{c}{R@5} & \multicolumn{1}{c}{R@10} & \multicolumn{1}{c|}{Time(s)} & \multicolumn{1}{c}{R@1} & \multicolumn{1}{c}{R@5} & \multicolumn{1}{c}{R@10} &\multicolumn{1}{c}{Time(s)}  \\ [-0.3ex] \midrule
    RaSa~\cite{bai2023rasa} &  \textbf{76.51} & {90.29} & \textbf{94.25} & 1168 &  {65.28} & {80.40} & {85.12} & 3871 &  \textbf{66.90} & \textbf{86.50} & \textbf{91.35} & 388 \\ 
    \ccol Ours &\ccol {75.28} &\ccol \textbf{90.42}  &\ccol {94.12}&\ccol16 &\ccol \textbf{65.76} &\ccol \textbf{81.39}  &\ccol \textbf{86.73} &\ccol 91  &\ccol {61.80} &\ccol {82.85}  &\ccol {89.45} &\ccol 5\\ \bottomrule
    \end{tabular}
    }
    \label{tab:comp_rasa}
    \vspace{0mm}
\end{table*}

\vspace{0mm}\section{More Ablation Studies}
\label{sec:more_ablation}

\noindent{\textbf{Ablation Study on Part Similarity Aggregations:}}
To demonstrate the effectiveness of our TDPA for similarity aggregation of part embeddings, we conduct comparative experiments with an alternative aggregation approach.
As demonstrated in Table~\ref{tab:more_ablation}(a), the cumulative sum is a similarity aggregation approach that simply sums the similarities across all part embeddings.
This approach inevitably leads to a decrease in performance due to the inclusion of non-informative elements, particularly affecting the R@1 metric.
Therefore, our TDPA method, by dynamically adjusting important part slots for each text query, significantly contributes to the enhancement of retrieval performance.

\noindent \textbf{{Benefits of adopting GRU in PSA block:}}
Within the PSA block, GRU facilitates learning-based decisions on the use of information from previous part slots for each update.
The ablation study in Table~\ref{tab:more_ablation}(b) empirically demonstrates that GRU contributes to the performance.

\noindent{\textbf{{Effect of the number of part slots:}}}
Table~\ref{tab:more_ablation}(c) shows that performance improves when there are enough part slots to segment the input data into distinct parts.
However, if more part slots are provided than the distinctive parts in the input data, it does not offer significant benefits.

\begin{table}[!t]
\centering
\vspace{-1.5mm}
\caption{
Ablation studies on CUHK-PEDES dataset. 
}
\resizebox{0.8 \textwidth}{!}{
\fontsize{9}{9}\selectfont
\begin{tabularx}{0.8 \textwidth}
{
  p{0.035\textwidth}
  p{0.35\textwidth}
  >{\centering\arraybackslash}X
  >{\centering\arraybackslash}X
  >{\centering\arraybackslash}X
  }
  \toprule
  & Methods & R@1 & R@5 & R@10 \\ \midrule
  \multirow{1}{*}{(a)}
  &Cumulative Sum & 73.96& 89.73 & 93.84\\ \midrule
  \multirow{1}{*}{(b)}
  & Ours w/o GRU & 74.33 & 89.49  & 93.52 \\ \midrule
  \multirow{3}{*}{(c)}
  & Ours w/ 4 part slots & 73.79 & 89.88  & 93.22 \\
  & Ours w/ 6 part slots & 75.02 & 89.93  & 94.23 \\
  & Ours w/ 10 part slots & 74.51 & 90.21  & 94.56 \\ \midrule
  & \ccol Ours & \ccol 75.28 & \ccol 90.42 & \ccol 94.12 \\
  \bottomrule
\end{tabularx}
}
\vspace{-1.0mm}
\label{tab:more_ablation}
\end{table}
\renewcommand{\thefigure}{\Alph{figure}}
\vspace{0mm}\section{More Qualitative Results}
\label{sec:more_qual}
Fig.~\ref{fig:supp_vis_attn_1} shows the more visualization of attention map $\bar{A}_{k}$ in $T$-th iteration of PSA block for both modality and TDPA weights $\mathbf{a}$ on CUHK-PEDES dataset.
Most of the presented results demonstrate that the part embeddings extracted by our part discovery module capture distinctive human parts.
Moreover, it shows the part embeddings extracted from the same part slot represent identical part although they are computed from different modalities. 
Furthermore, we can find the effectiveness of TDPA that enables adaptive part-based retrieval by focusing more on the distinctive human parts presented in the query text (Fig.~\ref{fig:supp_vis_attn_1}(a,b,c,e)).
Besides, TDPA addresses the limitation of slot attention; due to the mutually exclusive property of slot attention, a slot may draw attention on an irrelevant area if its corresponding part is not present.
For example, Part 4 semantically corresponds to the shoe areas, but since no shoe exists in Fig.~\ref{fig:supp_vis_attn_1}(b) and Fig.~\ref{fig:supp_vis_attn_1}(h), the associated slot instead attends to irrelevant regions.
However, to address this limitation, TDPA assigns low weights to Part 4 if shoes are not presented.

Also, we provide top-5 retrieval results of our method on three datasets are presented in Fig.~\ref{fig:supp_vis_top5_cuhk}, Fig.~\ref{fig:supp_vis_top5_icfg}, and Fig.~\ref{fig:supp_vis_top5_rstp}, respectively.
Most of the presented results illustrate that our method successfully retrieves the target individual.
We observe consistent retrieval results even in the presence of obstacles or variations in human poses.
In the retrieval results of false matches, the retrieved images are either very similar to other true matches (Fig.~\ref{fig:supp_vis_top5_cuhk}(e) and Fig.~\ref{fig:supp_vis_top5_rstp}(h)) or contain the distinctive human parts described in the text query (Fig.~\ref{fig:supp_vis_top5_icfg}(g) and Fig.~\ref{fig:supp_vis_top5_rstp}(f)).

\begin{figure*}[htbp]
    \centering
    \includegraphics[width=0.98\textwidth]{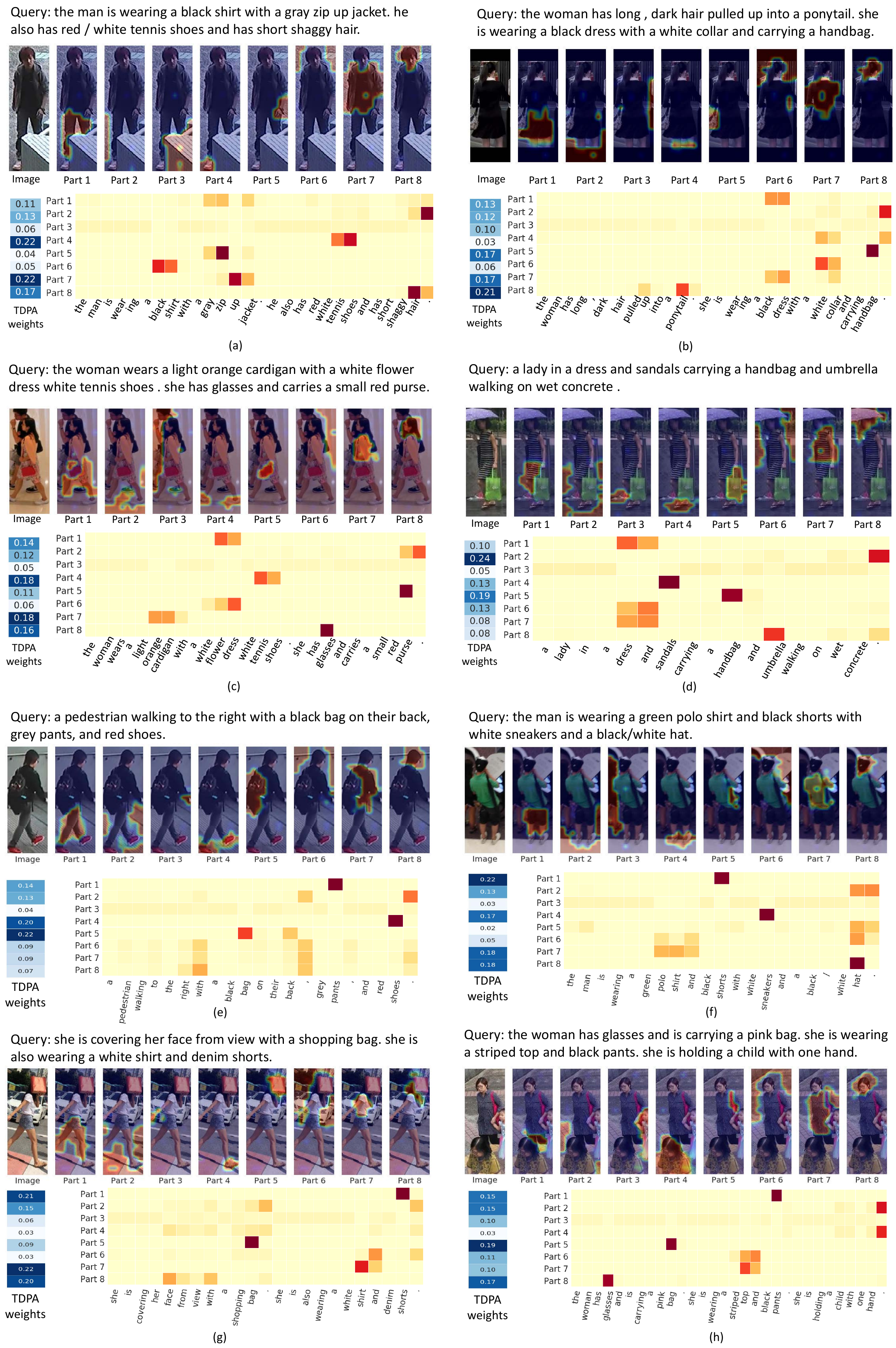}
    \vspace{-2mm}
    \caption{Visualization of each modality's attention map $\bar{A}_{k}$ in $T$-th iteration of PSA block and TDPA weights $\mathbf{a}$ on CUHK-PEDES dataset.
    }
    \vspace{-8.5mm}
    \label{fig:supp_vis_attn_1}
\end{figure*}

\begin{figure*}[htbp]
    \vspace{3mm}
    \centering
    \includegraphics[width=0.99\textwidth]{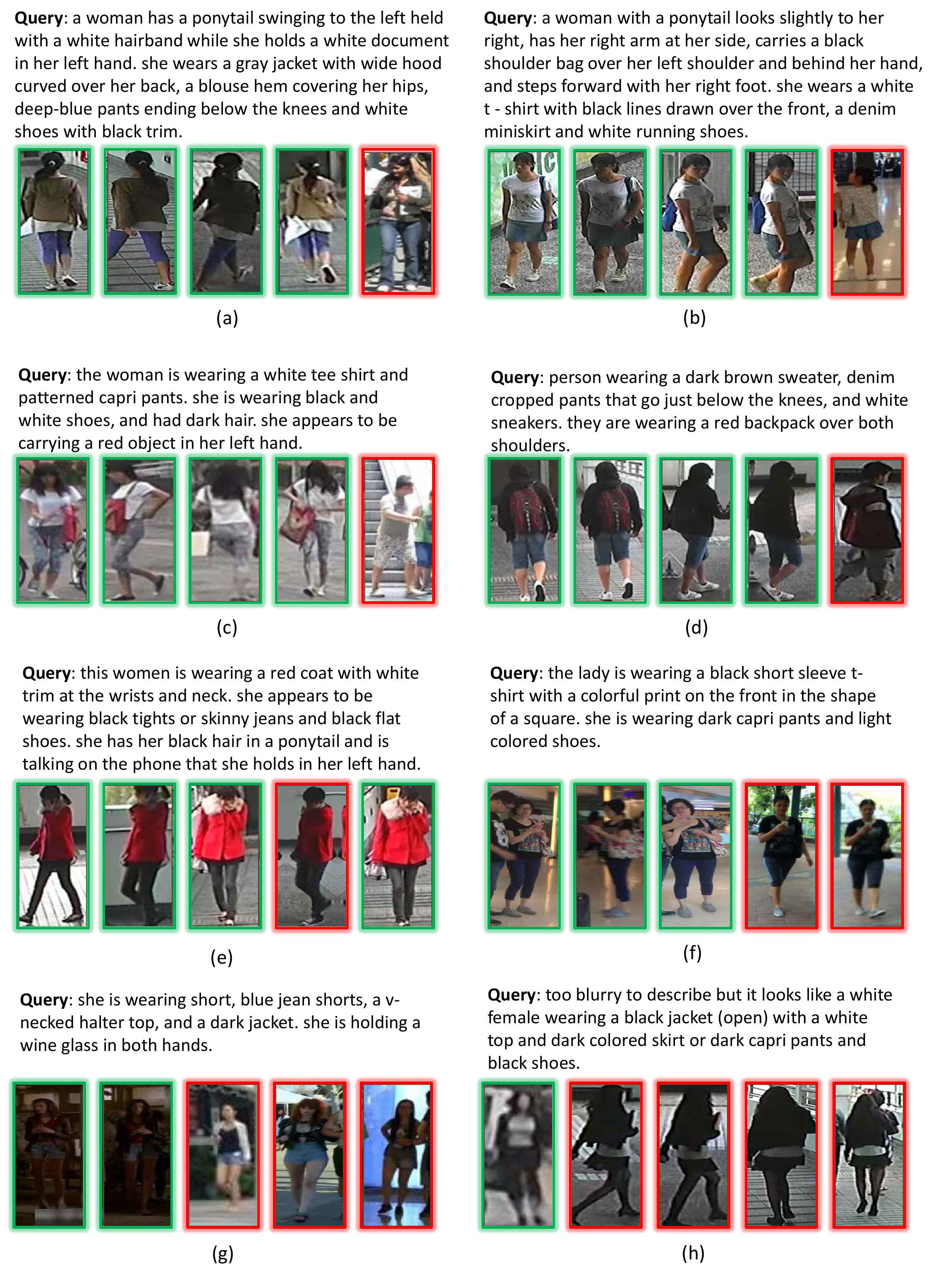}
    \caption{
    Top-5 retrieval results of our method on the CUHK-PEDES dataset.
    Images are sorted from left to right according to their ranks below each text query.
    Green and red boxes indicate true and false matches, respectively.
    }
    \label{fig:supp_vis_top5_cuhk}
\end{figure*}

\begin{figure*}[htbp]
    \vspace{3mm}
    \centering
    \includegraphics[width=0.99\textwidth]{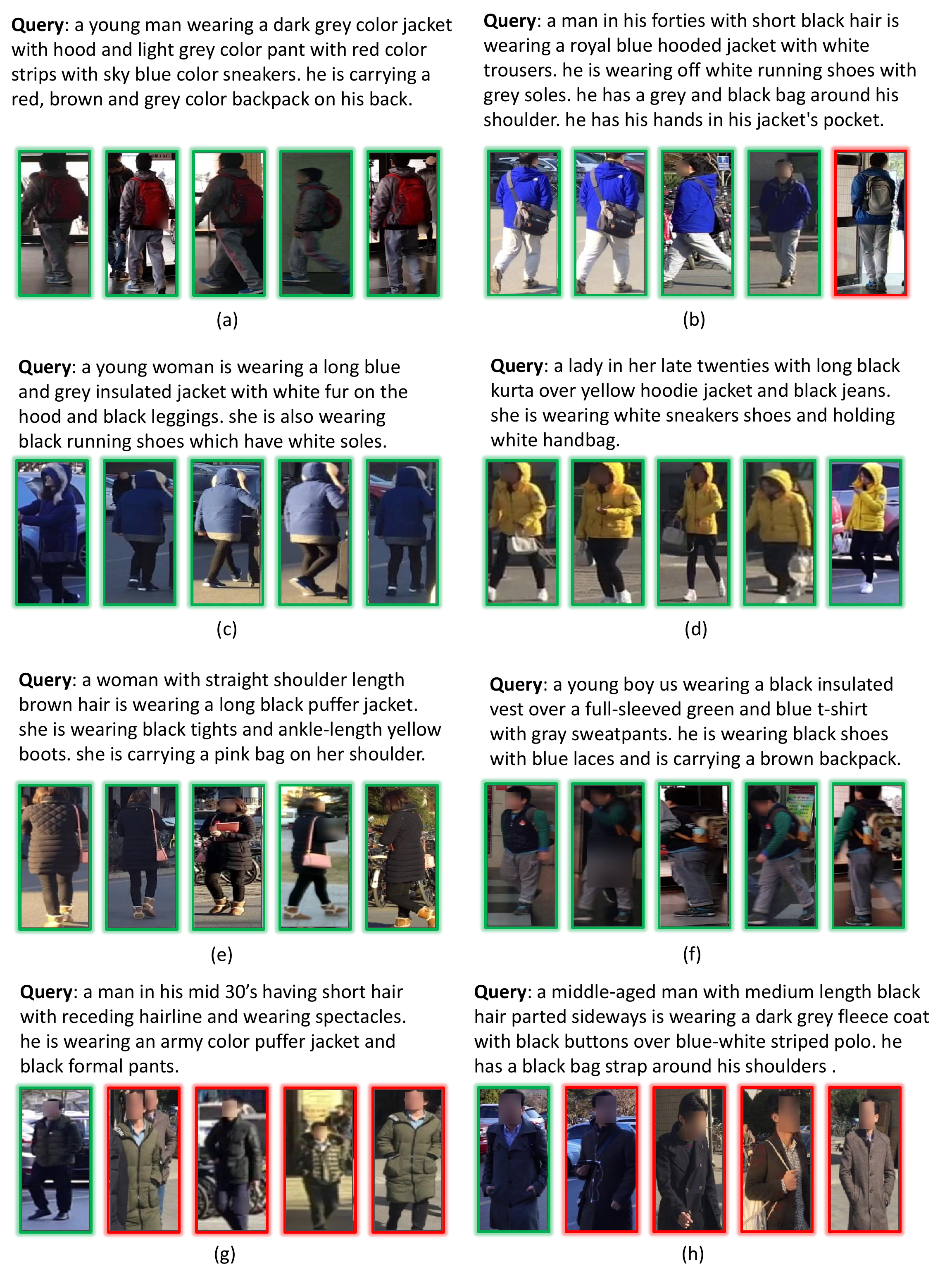}
    \caption{
    Top-5 retrieval results of our method on the ICFG-PEDES dataset.
    Images are sorted from left to right according to their ranks below each text query.
    Green and red boxes indicate true and false matches, respectively.
    }
    \label{fig:supp_vis_top5_icfg}
\end{figure*}

\begin{figure*}[htbp]
    \vspace{3mm}
    \centering
    \includegraphics[width=0.99\textwidth]{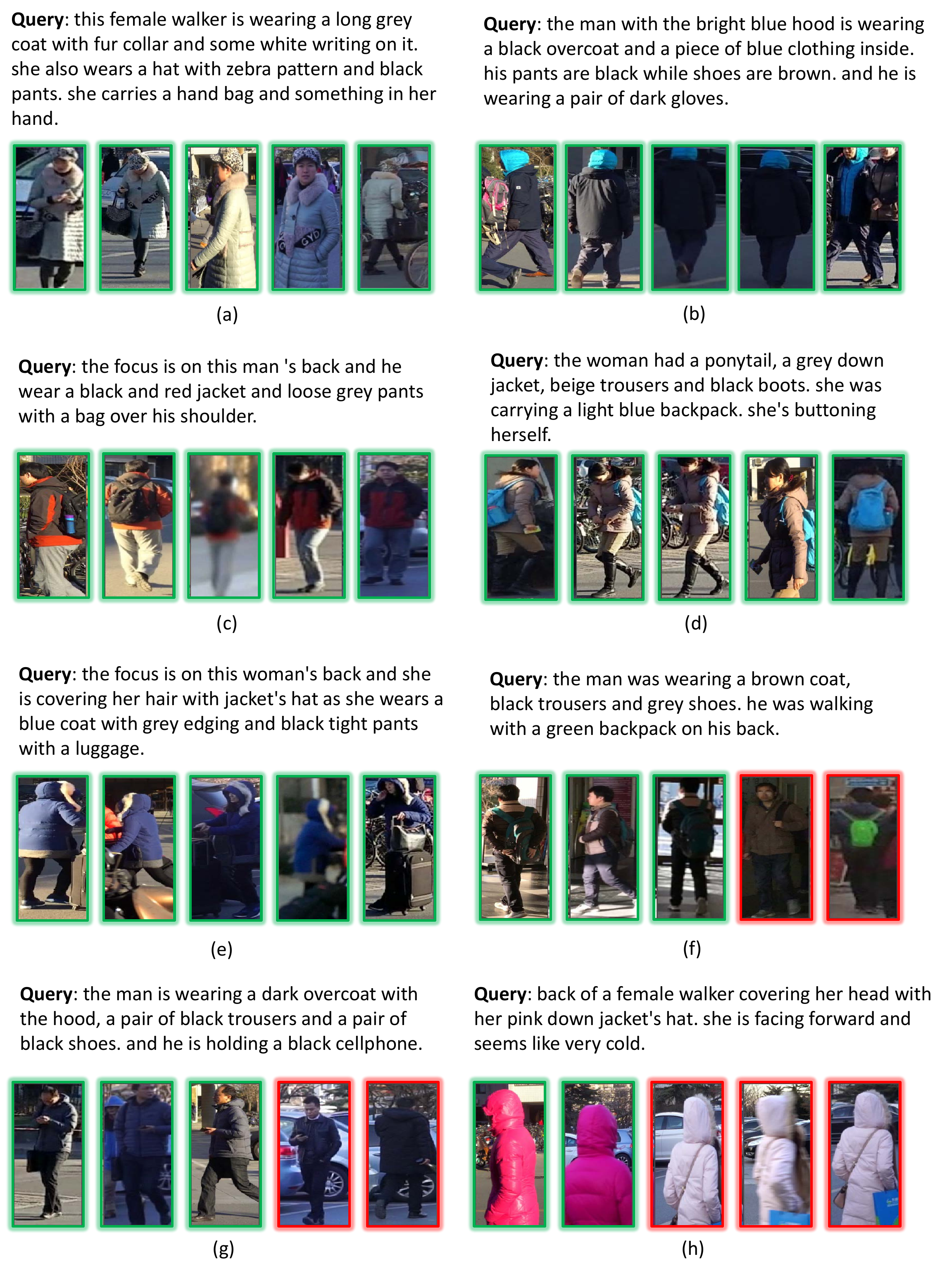}
    \caption{
    Top-5 retrieval results of our method on the RSTPReid dataset.
    Images are sorted from left to right according to their ranks below each text query.
    Green and red boxes indicate true and false matches, respectively.
    }
    \label{fig:supp_vis_top5_rstp}
\end{figure*}

\begin{figure*}[t!]
    \centering
    \includegraphics[width=1\textwidth]{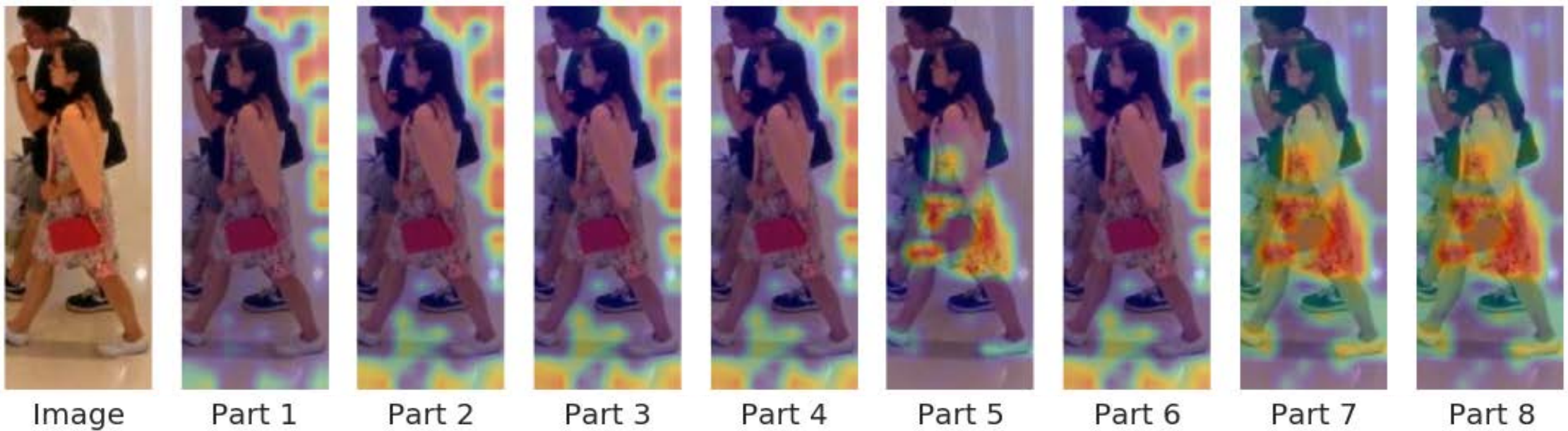}
    \caption{
    Visualization of part attentions with PAT applied to ours. 
    }
    \label{fig:supp_vis_pat}
\end{figure*}

\end{alphasection}



\end{document}